\RequirePackage{fix-cm}
\documentclass[twocolumn]{svjour3}          
\smartqed  
\usepackage[utf8]{inputenc}
\usepackage[T1]{fontenc}
\usepackage[english]{babel}
\usepackage{csquotes}
\usepackage[final,
            tracking=true,
            kerning=true]{microtype}
\usepackage{xspace}
\usepackage{etoolbox}
\usepackage{academicons}

\usepackage{amssymb}
\usepackage{amsmath}

\usepackage{graphicx}
\usepackage{booktabs}
\AtBeginEnvironment{tabular}{\scriptsize}

\usepackage{nth}
\usepackage{siunitx}
\usepackage{hyperref}
\hypersetup{
    colorlinks = true,
    linkcolor = {blue},
    urlcolor = {blue},
}
\addto\extrasenglish{

}

\newcommand*{\imgpix}{x}
\newcommand*{\cellcount}{l}
\newcommand*{\imgtype}{t}
\newcommand*{\typenat}{\mathfrak{n}}
\newcommand*{\typesynth}{\mathfrak{s}}
\DeclareMathOperator{\recocost}{Rec}
\DeclareMathOperator{\regrcost}{Reg}
\DeclareMathOperator{\regucost}{\mathcal{D}_{KL}}
\newcommand*{\weightparam}{C}
\DeclareMathOperator{\twincost}{Twin_{loss}}

\newcommand*{\dataset}[1]{\texttt{#1}}

\newcommand*{\dsnat}{\dataset{Nat}}
\newcommand*{\dssyn}{\dataset{Syn}}
\newcommand*{\dsbf}{\dataset{BF}}
\newcommand*{\dspc}{\dataset{PC}}
\newcommand*{\dstr}{\dataset{Tr}}
\newcommand*{\dste}{\dataset{Te}}
\newcommand*{\dsu}{\dataset{U}}
\newcommand*{\dsl}{\dataset{L}}

\newcommand*{\dsnatbf}{\dataset{Nat-BF}}
\newcommand*{\dsnatbfltr}{\dataset{Nat-BF-L-Tr}}
\newcommand*{\dsnatbfutr}{\dataset{Nat-BF-U-Tr}}
\newcommand*{\dsnatbflte}{\dataset{Nat-BF-L-Te}}
\newcommand*{\dsnatbfute}{\dataset{Nat-BF-U-Te}}

\newcommand*{\dsnatpc}{\dataset{Nat-PC}}
\newcommand*{\dsnatpcltr}{\dataset{Nat-PC-L-Tr}}
\newcommand*{\dsnatpcutr}{\dataset{Nat-PC-U-Tr}}
\newcommand*{\dsnatpclte}{\dataset{Nat-PC-L-Te}}
\newcommand*{\dsnatpcute}{\dataset{Nat-PC-U-Te}}

\newcommand*{\dssynbfltr}{\dataset{Syn-BF-L-Tr}}
\newcommand*{\dssynbflte}{\dataset{Syn-BF-L-Te}}

\newcommand*{\dssynpcltr}{\dataset{Syn-PC-L-Tr}}
\newcommand*{\dssynpclte}{\dataset{Syn-PC-L-Te}}

\newcommand*{\method}[1]{\texttt{#1}}

\newcommand*{\metcv}{\method{Watershed}}

\newcommand*{\metbit}{\method{BiT}}

\newcommand*{\metvae}{\method{Twin-VAE}}
\newcommand*{\metvaetran}{\method{Transfer Twin-VAE}}
\newcommand*{\dmetvaetran}{\method{double Transfer Twin-VAE}}
\newcommand*{\metvaemaxacc}{\method{Twin-VAE\textsubscript{max-acc}}}
\newcommand*{\metvaemindev}{\method{Twin-VAE\textsubscript{min-dev}}}

\newcommand*{\na}{n/a}
\newrobustcmd\best{\DeclareFontSeriesDefault[rm]{bf}{b}\bfseries}

\begin{document}

\title{Novel transfer learning schemes based on Siamese networks and synthetic data}


\author{Philip Kenneweg* \and Dominik Stallmann* \and Barbara Hammer \\
\begin{small}* with equal contribution \end{small}
}


\institute{Dominik Stallmann \at
              Machine Learning Group, Bielefeld University, Germany \\
              \email{dstallmann@techfak.uni-bielefeld.de} \\
           \and
           Philip Kenneweg \at
              Machine Learning Group, Bielefeld University, Germany \\
              \email{pkenneweg@techfak.uni-bielefeld.de} \\
             \and
           Barbara Hammer \at
              Machine Learning Group, Bielefeld University, Germany \\
              \email{bhammer@techfak.uni-bielefeld.de} \\
}

\date{} 

\authorrunning{Stallmann*, Kenneweg*, et al.}
\maketitle

\begin{abstract}
%
Transfer learning schemes based on deep networks which have been trained on huge image corpora offer state-of-the-art technologies in computer vision. 
Here, supervised and semi-supervised approaches constitute efficient technologies which work well with comparably small data sets. Yet, such applications are currently restricted to  application domains where suitable deep network models are readily available.
In this contribution, we address an important application area in the domain of biotechnology, the automatic analysis of CHO-K1 suspension growth in microfluidic single-cell cultivation, where data characteristics are very dissimilar to existing domains and trained deep networks cannot easily be adapted by classical transfer learning.
We propose a novel transfer learning scheme which expands a recently introduced Twin-VAE architecture, which is trained on realistic and synthetic data, and we modify its specialized training procedure to the transfer learning domain. In the specific domain, often only few to no labels exist and annotations are costly. We investigate a novel transfer learning strategy, which incorporates a simultaneous retraining on natural and synthetic data using an invariant shared representation as well as suitable  target variables, while it learns to handle unseen data from a different microscopy technology.
We show the superiority of the variation of our Twin-VAE architecture over the state-of-the-art transfer learning methodology in image processing as well as classical image processing technologies, which persists, even with strongly shortened training times and leads to satisfactory results in this domain.
The source code is available at \url{https://github.com/dstallmann/transfer\_learning\_twinvae}, works cross-platform, is open-source and free (MIT licensed) software. We make the data sets available at \url{https://pub.uni-bielefeld.de/record/2960030}.


\end{abstract}

\section{Introduction}
\label{intro}

Systematic single-cell studies of live cell imaging from microfluidic single-cell cultivation~(MSCC) works with high spatial and temporal resolution of cellular behavior.
So far, analysis of images like these has mostly been performed manually or is assisted by technological aiding systems, yet requiring human experts and therefore extensive human labor to create annotations of images; clearly, this procedure is not feasible in many cases, and it creates the need for different, more affordable and automated computer vision solutions~\cite{Theorell2019}.

The current state of the art for computer vision tasks and image processing that does not require human labor are convolutional deep neural networks~\cite{Ioannidou2017}. These are also  used extensively in the biomedical domain~\cite{Razzak2018}. Especially, approaches to track cells in images~\cite{Moen2019} have been an ongoing field of study in recent years. However, optimization for this task has proven to be a very cumbersome challenge which remains prone to errors. 

In recent years few shot learning has become a large research area. As an example interesting approaches constitute: ~\cite{transferlearning.xyz,NovelRec38:online,FewShotL77:online}. Popular surveys on this topic \cite{DeepAI-survey} categorize few shot learning into single modal and multi-modal learning. Here, single-modal learning can be further divided into Transfer Learning, Data Augmentation and Meta Learning. This work can be classified as a few shot learning approach that applies Data Augmentation and Transfer Learning for the specific use case of microscopy cell counting.

The proposed benchmark suite~\cite{Ulman2017} allows comparing different imaging technologies and extrapolation of the strengths and limitations of diverse methods for cell tracking, none of which are deemed as final solution for this task, even those with added interaction by bioimage analysis experts \cite{Berg2019} or distributed work of manual labeling~\cite{Hughes2018}.

In this contribution, we address a challenging task in biomedical image analysis by means of specific and adapted transfer learning technologies.
Related work in this field includes 
Brent et al.~\cite{Brent2018} which used transfer learning to predict microscope images between different imaging technologies, however without sufficient incorporation of the vast diversity of cell imagery and characteristics. The approach by Falk et al.~\cite{Falk2019} provides one of the few toolboxes for cell tracking, albeit adherent, rather than suspension cells.  It allows transfer learning based on given models and novel data, whereby data set enrichment technologies limit the number of required samples.

In contrast to already reported single-cell cultivation studies \cite{Carlo2006} and \cite{Kolnik2012}, where adherent growing cell lines are the focus of investigation, we address  the scenario of more complex suspension cells, with their circular basic shape but ever-changing contour due to vesicle secretion and additional challenges like cell movement and floating within the experiment chamber, which renders analysis tools of adherent cells deficient. These cells growing in suspension comes with different and challenging obstacles to achieve automation of analysis, which will be described in \autoref{sec:data}.

Siamese networks have been used for a variety of tasks as they can help facilitate few shot learning or clustering of the data space by generalizing from unlabeled data. This is done in \cite{muller-etal-2022-shot} for text data and in \cite{10.3389/fcell.2021.767897} for genome sequencing. These presented architecture are however specific to their domains and not applicable to image processing. Furthermore, we extend the classical siamese network by including transfer learning and automatic data augmentation.

In this work, we want to make use of a network trained on one microscopic image type and adapt it to provide sufficiently accurate cell counting for a different microscopy technology, where no trained network exists due to the lack of annotated data. We particularly focus on mitigating human labor for annotations.
Our previously introduced deep twin auto-encoder architecture \metvae{} \cite{TwinVAE} is trained on data stemming from one imaging modality and thereafter transferred to the similar yet different domain of the other microscopy technology.
This training procedure greatly reduces the need for natural, labeled data, by using synthetic, auxiliary training data, for which the ground truth is known and which is easy to obtain in this setting, since the \metvae{} does not require the images to be rendered realistically in every regard, such as morphological details.

In the following, we will first describe the specific application domain from biotechnology, the underlying machine learning challenge, and the deep Siamese network architecture which will be used for transfer learning. Afterwards, we elaborate the details of the proposed transfer learning scheme, as well as perform an analysis of how the unique architecture used affects the transfer learning procedure. Its performance is evaluated for real data sets and using ablation studies, as well as comparison to state-of-the-art alternatives and baselines. A discussion concludes the contribution.

The application area in question is a prime example of a domain, where the state-of-the-art Image processing techniques do not work sufficiently well due to very little texture and other visual characteristics of the images, described in \autoref{sec:data}.
In addition, there exists no Deep Learning Models which easily and efficiently solve the task, as shown in \cite{TwinVAE} by comparing to EfficientNet \cite{Tan2019}, and Watershed methods \cite{Watershed2013} and shown here later by comparing to BigTransfer \cite{DBLP:journals/corr/abs-1912-11370} and our previous work.
The emergence of more data in such specialized domains like this makes it important to provide an easy-to-use system which has a high performance and enables automation of processes involving this kind of data.
\newline
Thus, the contribution and novelty of our work is as follows:
\begin{itemize}
    \item We improved performance and lowered computational complexity (outperforming the original work \cite{TwinVAE})
    \item We build an efficient transfer pipeline and showed on two microscopy datasets empirically that it outperformed a variety of methods, including state-of-the-Art Image processing. 
    \item By performing extensive ablations studies we gained insight into
    which parts of the architecture contains representations which are beneficial for the transfer learning ability of the network. Thus, we are contributing to the debate how deep neural networks represent information. \cite{natureNNunderstanding} 
\end{itemize}

\section{Material and Methods}

\label{sec:data}
\subsection{MSCC and live cell imaging data}

\begin{figure}[t!]
\centering
\includegraphics[width=\linewidth]{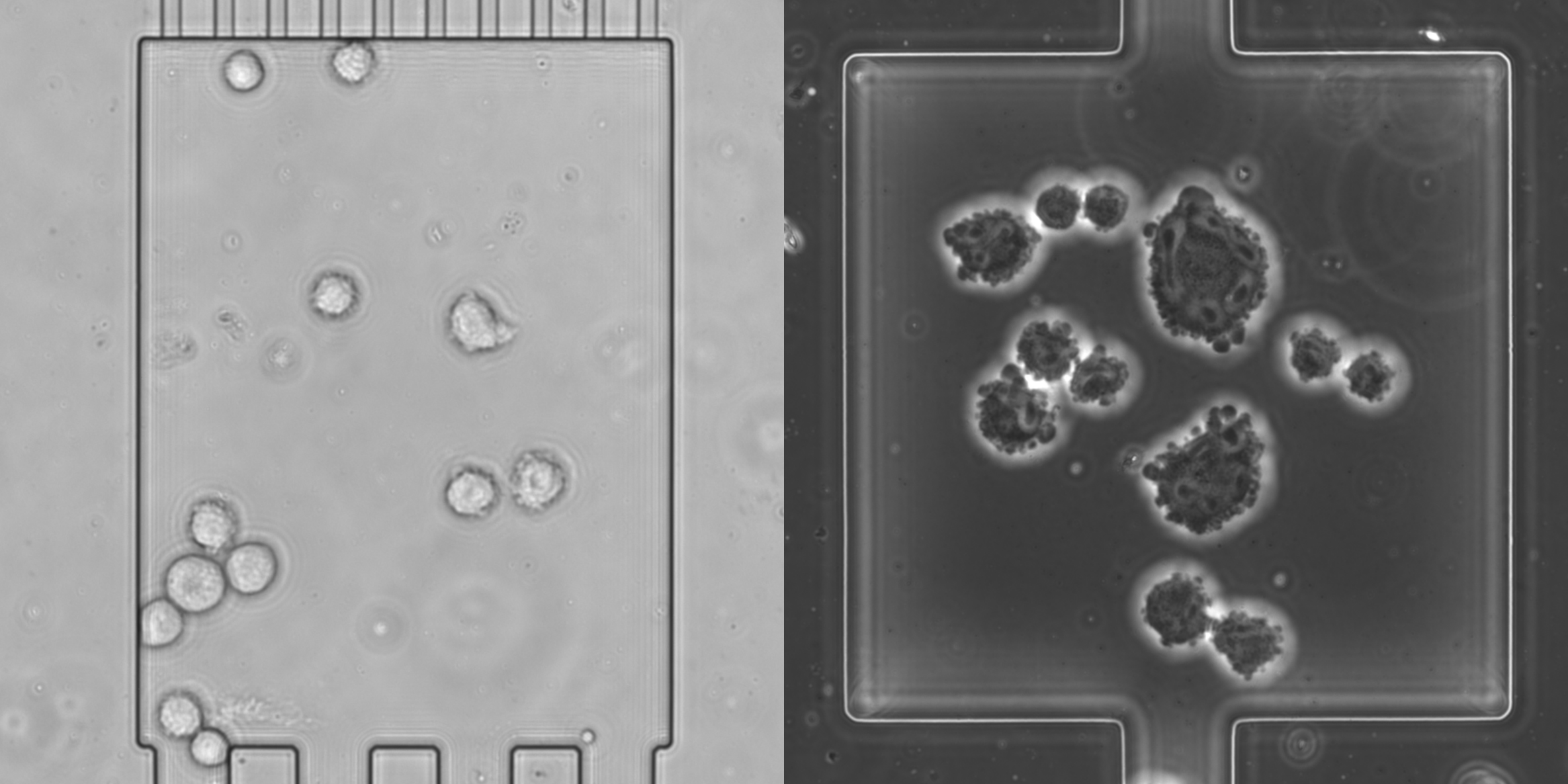}
\caption{\ Samples of data from the two microscopy technologies. Bright-field microscopy (left), phase-contrast microscopy (right). 
The data has been preprocessed in the form of chip drift removal and orientational stabilization (translational and rotational) and a crop to square the images that allows further cropping by data augmentation techniques.
The cell counting module has to differentiate between the cells, smudges, chamber and background.}\label{fig:sample}
\end{figure}

The image data which is used in this study was obtained by single-cell cultivation of mammalian suspension cells as shown before~\cite{Schmitz2020}. CHO-K1 cells were cultivated in polydimethylsiloxane (PDMS)-glass-chips and constantly provided with nutrients by perfusion of the microfluidic device. 
The goal of an automated analysis of such data is an automated extraction of important parameters of the observed dynamics, such as cell growth. Since many important parameters can be estimated based on the number of cells at a specific time point, the number of cells constitutes a key quantity and are taken as target labels.
The data used in this work consists of multiple parts, characterized 1) by the according microscopy technologies, bright-field microscopy and phase-contrast microscopy, 2) by the type of data, natural or synthetic, i.e., the original data or data which are generated and added to the original one within the learning pipeline, as described later, 3) by existence of a label and 4) the usage of that data for training or testing. Example images of both modalities are shown in \autoref{fig:sample}. 

\autoref{Tab:01} shows an overview statistic of all data sets. The \texttt{Nat} set contains the aforementioned natural images, \texttt{Syn} the synthetic ones. The \texttt{BF} tag declares bright-field microscopy images, consisting of \num{12}~experiment scenarios with \num{956} overall images, of which a label (i.e.\ cell count) exists for \SI{7.5}{\percent} of the training images.
The \texttt{PC} tag is denoting phase-contrast microscopy images which consist of \num{37}~experiment scenarios, accumulating to \num{3976}~used images with a labeling rate of \SI{6.2}{\percent} for the training data.
The labels were created by hand in a nearly regular interval over the experiment scenarios for all natural data sets, however images were removed beforehand, if they had more than \num{30}~cells, since the expected outcome of the cultivation experiment is already determined at this point.

\begin{figure}[t]
\centering
\includegraphics[width=\linewidth]{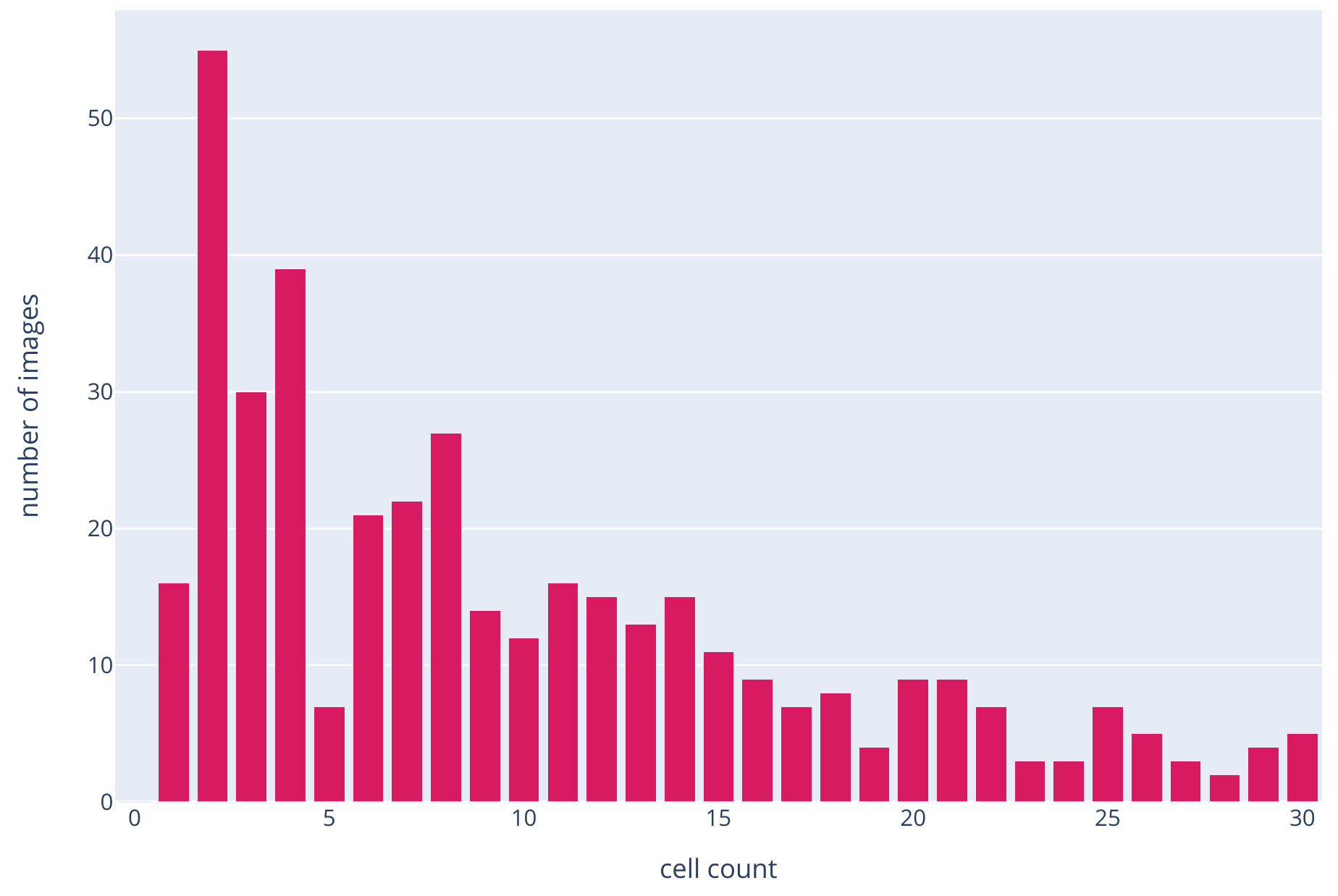}
\caption{Visualization of distribution of images by cell count for the merged data sets \dsnatpclte{} and \dsnatpcute{}. We discard data with higher cells counts than \num{30}, because they are irrelevant for the cultivation experiments.}\label{fig:distribution}
\end{figure}

In the upcoming analysis, we focus on the transfer from the larger data set \dsnatpc{} to the smaller set \dsnatbf{}, since this is the common way to apply transfer learning. The phase-contrast imagery also contains more variation of the biological processes, which makes phase-contrast microscopy arguably more popular than bright-field microscopy. Our experiments also contains transfers from bright-field microscopy to phase-contrast microscopy to show the robustness of the technique.
\autoref{fig:distribution} shows the distribution of images against the cell counts in them for \dsnatpc{}. A clear trend towards images with low cell counts can be seen. 

This can be taken into account for optimization of the transfer learning methodology, since it can be assumed that data of this type has a similar distribution, particularly in the light of exponential growth rates and the presence of failing cultivations.
Most of the labeled images from (\dataset{L-Te}) are used for testing the cell count prediction and the reconstruction, rather than used during training (\dataset{L-Tr}). This is done, because we focus on a method that reliably works on small amounts of labeled data.
Further unlabeled test data (\dataset{U-Te}) is used to evaluate the reconstruction only, since (\dataset{L-Te}) remains too small to include a broad overview of the different chamber situations (clumping, overlapping, escaping etc.) to be confident about the stability of the performance for a convolutional network.

\begin{table}[!b]
\caption{Overview of data sets used. The \dataset{Nat} tag indicates natural data, \dataset{Syn} represents synthetic data. \dataset{BF} classifies the bright-field images, \dataset{PC} the phase-contrast ones. \dataset{L} and \dataset{U} marks labeled and unlabeled data and lastly \dataset{Tr} and \dataset{Te} separate the data into training and test data.}
\label{Tab:01}
{
\centering
\setlength{\tabcolsep}{0.5em} 
\begin{tabular}{@{}cccccS[table-format=4.0]@{}}\toprule
Abbreviation & Type      & Technique      & Label & Usage      & {Size} \\\midrule
\dsnatbfltr  & natural   & bright-field   & yes     & Training &    281 \\
\dsnatbfutr  & natural   & bright-field   & no      & Training &    2188 \\
\dsnatbflte  & natural   & bright-field   & yes     & Testing  &    290 \\
\dsnatbfute  & natural   & bright-field   & no      & Testing  &    224 \\
\dsnatpcltr  & natural   & phase-contrast & yes     & Training &    209 \\
\dsnatpcutr  & natural   & phase-contrast & no      & Training &    2943 \\
\dsnatpclte  & natural   & phase-contrast & yes     & Testing  &    398 \\
\dsnatpcute  & natural   & phase-contrast & no      & Testing  &    394 \\
\dssynbfltr  & synthetic & bright-field   & yes     & Training &    2469 \\
\dssynbflte  & synthetic & bright-field   & yes     & Testing  &    514 \\
\dssynpcltr  & synthetic & phase-contrast & yes     & Training &    3152 \\
\dssynpclte  & synthetic & phase-contrast & yes     & Testing  &    792 \\\bottomrule
\end{tabular}
}
\end{table}

\subsection{Synthetic data}
As our task is reliable cell counting for suspension cell microscopic images and given data is often limited and with only few manual annotations, retraining a deep neural network for every new set of data is inadequate and delivers deficient accuracies for the task. To overcome this limitation, we transfer a trained model which achieves high accuracies on its original task to the newly presented task.

Since the learning methodology is semi-supervised, our formerly introduced Twin-VAE \cite{TwinVAE} will be used as a basis to propose a novel transfer learning method to mitigate the aforementioned complications. Here, synthetic data is used as an auxiliary training set \dssyn{} which is also used for transfer learning. We evolve on the Siamese architecture, which inherently solves the task of abstraction from the synthetic nature of data set enrichtments.

\begin{figure}[t!]
\centering
\includegraphics[width=\linewidth]{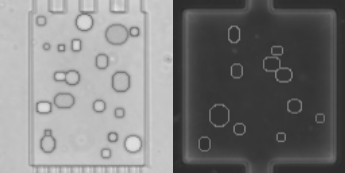}
\caption{Random samples of synthetic data from the sets \dssynbfltr\ and \dssynpcltr. Sample of bright-field microscopy on the left and of phase-contrast microscopy on the right. The images are dimensionless and therefore do not show scale bars. The theoretical cell sizes are identical to natural data cells, but is of no importance for the work.}\label{fig:sample_syn}
\end{figure}

The synthetically generated data is visually simplified (constant background, ellipsoidal cells) to allow the loss construction to focus on the regression task rather than the intricate reconstruction of arbitrary visual cell membranes and organelles. This is done by drawing cells as ellipsoids, varying some attributes like their brightness, size and blurriness of edges. For further detail, see the original Paper \cite{TwinVAE}. Reconstructions of real appearances from synthetic data, while interesting to suggesting inherent stability, are not of importance for a high accuracy on the task. Ground truth labels are known for synthetic data, because it is based on pre-defined geometric style modeling, neglecting texture and complex morphology. Thereby, geometric heterogeneity of this data is simplified compared to real data, examples of which can be seen in \autoref{fig:sample_syn}.

Synthetic data allows for creation of a large variety of independent image samples that are correlated but not identical to the appearance of natural data. Unlike popular data set enrichment technologies, the amount of data can freely be determined since it is independent of the amount of real data, and representatives of any type of underlying label can easily be generated.
We show that our \metvae{} architecture is successfully trained and improved in accuracy like this in \autoref{sec:model}.

\autoref{Tab:01} lists the synthetic data as \dssyn{}, concatenated by the microscopy technology category \dsbf{} or \dspc{} accordingly. The \dsu{} or \dsl{} declaration tells if the data is labeled and the table separates them between training (\dstr{}) and test (\dste{}) images.
The cell distribution in these images was chosen to be close to that of the \dsnat{} data sets for the sake of resemblance and because of the criticality of correct cell counts on low cell count imagery.
Synthetic images are generated with seed consistency, i.e. in such a manner that ensures reproducibility and can be generated in arbitrary amounts, however larger amounts of synthetic data will increase training time nearly linearly, while improving performance with diminishing returns as will be shown in \autoref{sec:evaluation}. For the sake of computational time, the default \dssyn{} set sizes roughly match the corresponding number of natural images. However, sets with different amounts have also been created for accuracy comparisons (\autoref{fig:syndata}).
The background is created by the mean value of the entire natural training data set showing less than \num{5} cells to assure high visibility of empty background and to remove the smudges that appear in only a few experiment scenarios.
The working resolution the synthetic data is generated in is \num{128}~by \num{128}~pixels, matching the resolution of the architecture's input size described in \autoref{sec:model} and their file size is about \SI{8}{MB} per \num{1000}~images.

The virtual generator is highly adjustable, creating images with a given distribution of cell counts,  overlapping of cells, variations of  the brightness of the cell's inner organelles, their membrane silhouette and the background.
More complex visual fidelity of natural data such as ongoing cell divisions can also be mirrored by a combination of these mechanics, e.g., by creating a small overlap together with more noisy cell borders. Smudges, as in \autoref{fig:sample}, have not been inserted, since they are an interference factor and likely only hinder the training process.
The cells' shape has been simplified to deformed ellipsis to roughly match the shape of the natural cells. Noise, individual luminance per cell, and multiplication with Gaussian filters of random strengths have been added to increase the variety of cells in the data.
We ensure easy adjustability of the generation mechanism to natural cells in other data sets, that have different shape characteristics.

\subsection{Network Architecture \& Training}
\label{sec:model}
\subsubsection{Siamese Architecture}
We use a novel deep Siamese twin architecture that separates the input data for training depending on its origin, thus circumventing the problem of differences in appearance of synthetic and natural images. 
This approach requires that the architecture creates a tightly coupled shared inner representation of the different data sources to achieve low training losses and good generalization ability for semi-supervised setups.

\begin{figure*}[t]
\centering
\includegraphics[width=1\linewidth]{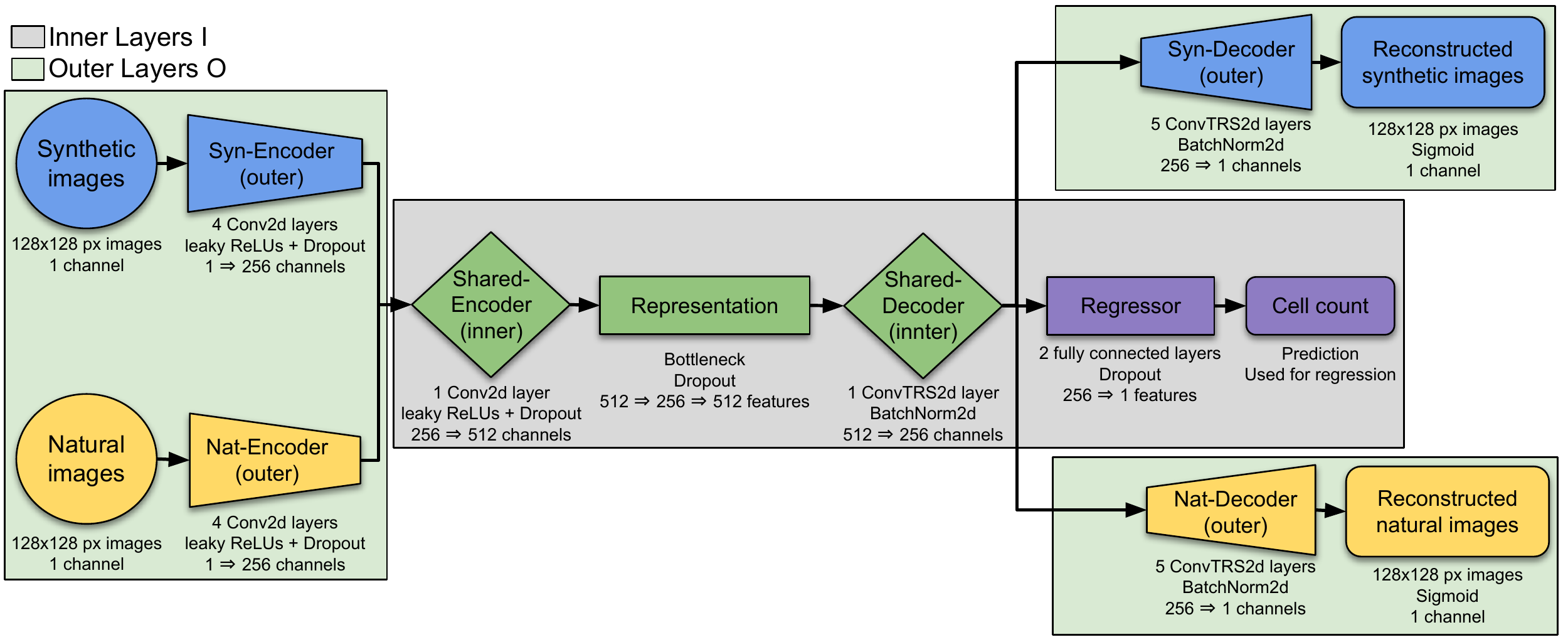}
\caption{Visualization of the \metvae{} architecture.
The blue elements handle synthetic data, while the yellow elements handle natural data.
The green elements are shared between the two VAEs and contain the inner representation of the cell imagery, while the purple elements result in an estimation of the cell count.}\label{fig:architecture}
\end{figure*}

For this, two identical variational autoencoders (VAE) are created for the two data sets. They share the weights of their last encoding layer, the first decoding layer and the small hidden layer in-between (see \autoref{fig:architecture}).
VAEs constitute a state-of-the-art solution for generalized few-shot learning~\cite{Schonfeld2019} and weight-sharing has been used to reduce neural network sizes and to improve test performance beforehand~\cite{Karen2017}.

Specifically, in our setup, one of the VAEs works on synthetic data only (\textit{VAE-syn}), while the other one uses natural data only (\textit{VAE-nat}).
The non-shared outer layers account for the different visual characteristics of synthetic and natural data, while the shared inner layers are enforced to create a common representation of relevant image characteristics. By adding a two-layer deep fully connected neural network regression model for the cell counting task, the architecture works in a supervised manner for data for which the label is known, based on the shared representation of the VAEs.
Cell detection by regression has been shown to work well for other (less demanding) tasks~\cite{Yuanpu2015,Weidi2018}.
Our architecture therefore addresses two objectives simultaneously:\\
(i) Mostly unsupervised encoding and decoding of natural and synthetic input images using a shared representation.\\
(ii) Supervised counting of the cells for both natural and synthetic images.

\subsubsection{Loss}
Given an input image \(\imgpix\) of pixels, a label (i.e., cell count) \(\cellcount\) between \num{1} and \num{30} and a type \(\imgtype \in \{\typenat, \typesynth\}\), representing the fact whether the image is natural or synthetic, we obtain a reconstruction loss \(\recocost(\imgpix)\) of the VAE, a regression loss \(\regrcost(\imgpix, \cellcount)\) of the task at hand such as cell counting, and a distributional regularization loss \(\regucost\), which aims for a homogeneous representation of synthetic and real data in the embedding space of the VAE.
We combine these losses to form our twin loss \(\twincost(\imgpix, \cellcount, \imgtype)\) with weighting factors \(\weightparam_{\recocost}^{\imgtype}\), \(\weightparam_{\regrcost}^{\imgtype, \cellcount}\), and \(\weightparam_{\regucost}^{\imgtype}\), respectively, which allows us to balance image reconstruction fidelity (\(\weightparam_{\recocost}^{\imgtype}\)), regression performance (\(\weightparam_{\regrcost}^{\imgtype, \cellcount}\)) and distributional stability (\(\weightparam_{\regucost}^{\imgtype}\)) and therefore to maximize the impact of regression errors on the loss. Furthermore, it allows us to gracefully handle input images without known cell counts by setting \(\weightparam_{\regrcost}^{\imgtype, \cellcount}\) to zero:
\begin{equation}\label{eq:twinloss}
\begin{split}
    \twincost(\imgpix, \cellcount, \imgtype) = \\
    \weightparam_{\recocost}^{\imgtype} \cdot \recocost(\imgpix)
    + \weightparam_{\regrcost}^{\imgtype, \cellcount} \cdot \regrcost(\imgpix, \cellcount)
    + \weightparam_{\regucost}^{\imgtype} \cdot \regucost(\imgpix)
\end{split}
\end{equation}

During our experiments, the mean-squared error (MSE) \(||x-d(x)||^2\), where \(d(x)\) is the reconstruction of the input image \(x\) and \(||l-r(x)||^2\), where \(r(x)\) is the estimated cell count, yielded the best results respectively, when used as \(\recocost(\imgpix)\) and \(\regrcost(\imgpix, \cellcount)\) for training on phase-contrast data, and as \(\regrcost(\imgpix, \cellcount)\) for bright-field data. However, for bright-field data the binary cross entropy (BCE) \(-l\cdot log(r(x))+(1-l)\cdot log(1-r(x))\)
turned out to be the superior choice for \(\recocost(\imgpix)\) and was often resulting in just slightly worse results than the MSE for phase-contrast data.
The \(\regucost\) is applied as the Kullback-Leibler divergence (KLD) of the standard VAE model (\cite{Kingma2013}) and is obligatory to enforce generation of latent vectors with sufficient similarity to a normal distribution.
The weighting factors \(\weightparam_{\recocost}^{\imgtype}\), \(\weightparam_{\regrcost}^{\imgtype, \cellcount}\), and \(\weightparam_{\regucost}^{\imgtype}\) are carefully chosen for training, punishing incorrect cell count predictions especially on natural data, while relaxing the importance of visual reconstruction.
Details on this are provided in the following section.

\subsubsection{Neural Network Structure}
\label{subsec:nn}
The outer, non-shared part of the encoder is composed of four two-dimensional convolutional layers with kernel size \num{5} and a stride of \num{2}, initialized with an orthogonal basis~\cite{Saxe2014}.
Inbetween the layers, leaky rectified linear units (ReLU) with a leakiness of \num{0.2} are activated, together with a dropout of \num{0.1}.
Channel amounts used for the convolutions are in order: \num{32}, \num{64}, \num{128} and \num{256} for the encoders.
The inner, shared part of the encoder consists of an additional two-dimensional convolutional layer with identical properties and \num{512} channels.
The layer is followed by the bottleneck, consisting of three fully connected layers of sizes \num{512}, \num{256} and \num{512} again, each with a dropout of \num{0.1}.
The inner, shared part of the decoder has \num{256}~channels, contains a two-dimensional transposed convolutional operator layer with identical kernel size and stride as in the encoder, and is followed by a batch normalization over a four-dimensional input and a leaky ReLU with the same leakiness.
The outer, non-shared part of the decoder consists of five layers of kernel sizes \num{5}, \num{5}, \num{5}, \num{2}, \num{6}, following the convention of a small penultimate followed by a bigger last layer, keeping the stride of \num{2} except for the fourth layer using a stride of \num{1}, the same leaky ReLUs and a sigmoidal activation function at the end.
Additionally, a branch of fully connected neurons for the regressor consisting of two layers of sizes \num{256} and \num{128} is being fed by the output of the shared part of the decoder, uses linear layers and a constant dropout of \num{0.1}.

The architecture is using the Adam optimizer for phase-contrast microscopy data, and the rectified Adam (RAdam)~\cite{Radam2020} optimizer for bright-field data.
The combination of the decoder loss factor \(\weightparam_{\recocost} = 100\), the regressor loss factor \(\weightparam_{\regrcost}=3\) and the KLD factor \(\weightparam_{\regucost}=2\) yields the best results for phase-contrast data.
For the BCE, the decoder loss factor is not constant, but decays over time with a rate of \num{3e-5} per epoch, since the BCE does not decrease significantly within the training process, but needs to decrease over time to amplify the importance of low regression losses \(\regrcost(\imgpix, \cellcount)\).

While it seems counter-intuitive that \(\weightparam_{\recocost}\) is bigger than \(\weightparam_{\regrcost}\) and \(\weightparam_{\regucost}\), it is caused by the MSE for pixel data getting very small on normalized images. KLD is supposed to  stay relatively small. While it is required to enhance the quality of the distributions, it should not impact the training of cell predictions and image reconstructions too much by unfortunate sampling from the latent vector, however it has to be impactful enough to enforce natural and synthetic data into similar representations in the inner layers.

Since training is done over thousands of epochs, a soft weight decay of \num{1e-5} per epoch is added, combined with a fixed learning rate of \num{1.3e-4}. A delayed start for the regressor is used to allow for pure image reconstructions to contain meaningful images, ensuring the representation of information of existent cells in the representation before the regressor has to extract that information. A delay of \num{100}~epochs has been used to achieve the results presented in \autoref{sec:evaluation}.

A batch size of \num{128} for the phase-contrast images and \num{64} for the bright-field images works best, and the training runs for up to \num{50.000} epochs, unless early stopping conditions abort it.


\subsubsection{Data Augmentation}
\label{subsec:da}
To maximize the use of the limited amounts of natural data, multiple data augmentation techniques are combined and applied to the data.
Randomly occurring horizontal and vertical flips, possibly combined with a random crop of the image of scale \num{0.9} combined with a resize to its original, meaning the images get randomly cropped to \num{115} pixels in width and height, then scaled back to \num{128} pixels.

The crop adds difficulty to the cell detection process by partially cropping cells out, however it proved helpful as long as the crop is not too strict and cuts away cells completely.
Then, a \num{90} degree rotation is applied at random and a zero-centered noise map is generated and added to the image with a small amplitude factor.
Additionally, small rotations of \num{0} to \num{5} degrees are added before the crop, to spread cell occurrence even more. The crop will then mostly remove the undefined parts of the image, that are created when rotating non-circular images.

\subsection{Image Reconstruction}
Although our goal is automatic counting of cells, our loss from \autoref{eq:twinloss} includes a term for image reconstruction. The reasoning behind this decision is that analysis of the reconstruction abilities of \metvae{} is only possible with this loss. Furthermore, the loss enables us to check if the learned shared representation is meaningful by checking the correlation between the visual existence of cells in the image reconstructions and the actual cell count.

During training of \metvae{}, the natural input images are first processed by a specialized encoder, followed by a shared encoder and decoder of the two twins, and finally reconstructed by a specialized decoder (see \autoref{fig:architecture}). Synthetic data is handled equivalently.
The learned inner representation must be shared and similar between the two types of data for 1) the regression to work as intended and 2) the cell counting in natural images to benefit from synthetic data as much as possible.
Verification of this is done by encoding a natural image with the appropriate encoder but performing the reconstruction with the decoder that is intended and trained for synthetic images and vice versa. In the following, we demonstrate exactly this.

\begin{figure}
\centering
\includegraphics[width=\linewidth]{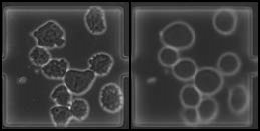}
\caption{Example of a perfect synthetic-looking reconstruction (right) of a natural image (left) from \dsnatpclte.
The cell counts match exactly and the position as well as size of cells are preserved.
While the smudge on the left is recreated visually, it does not lead the regressional part of the \metvae{} to a wrong cell count.}
\label{fig:optimal_pc}
\end{figure}

\begin{figure}
\centering
\includegraphics[width=\linewidth]{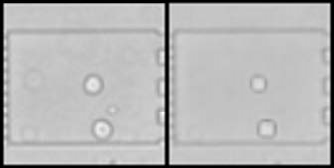}
\caption{Example of a perfect synthetic-looking reconstruction (right) of a natural image (left) from \dsnatbflte.
The cell counts match exactly and the position as well as size of cells are preserved. For this data set, where smudges are more faint, they don't get reconstructed usually.}
\label{fig:optimal_bf}
\end{figure}

In \autoref{fig:optimal_pc} and \autoref{fig:optimal_bf} we show examples of perfect translations, where a natural image is encoded and subsequently decoded as a synthetic image.
The cell count is unchanged, the cell prediction matches the actual existence of cells and the position and size of cells are also retained, while the overall appearance is simplified, however \metvae{} has learned to remove noise and condense the information down to what is required and helpful to count cells.


\begin{figure}
\centering
\includegraphics[width=\linewidth]{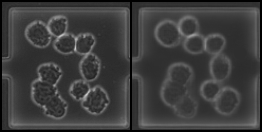}
\caption{Example of a faulty synthetic reconstruction (right) of a natural image (left) from \dsnatpclte.
The human expert determined the cell count to be \num{10}, the prediction differs by one. The reconstruction shows a merge of the top two cells in the bottom-left triple of cells. The two cells clump together in such a way, that there is almost no visual indication of a border between them, especially missing the usual bright boundary around cells that can be seen around the rest of the cells.}
\label{fig:suboptimal}
\end{figure}

\begin{figure}
\centering
\includegraphics[width=\linewidth]{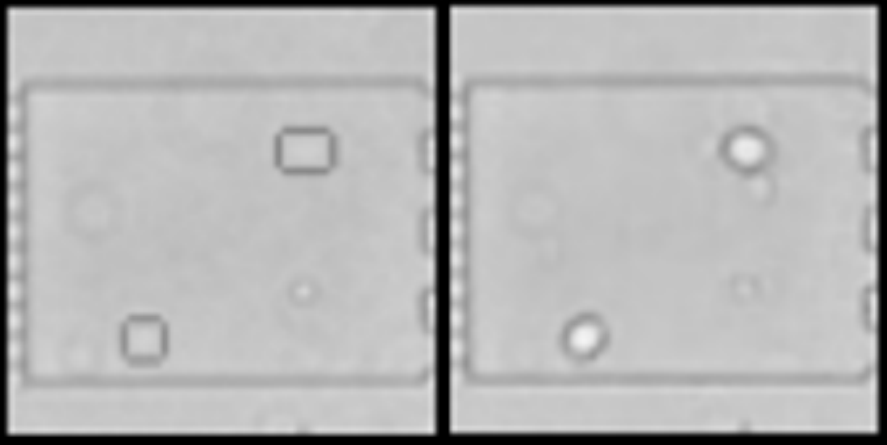}
\caption{Example of an accurate, natural-looking reconstruction (right) of a synthetic image (left) from \dssynbfltr. Cell counts match exactly, position and size of the cells are preserved. While these conversions are not mandatory for the transfer process, they ensure representational consistency on a visually comprehensible level.}
\label{fig:syntonat}
\end{figure}

Even when \metvae{} does not translate an image perfectly,
the reconstruction can be useful to understand where an error occurs.
In \autoref{fig:suboptimal} we show an example where two cells that are very close together are interpreted and reconstructed as a single cell.
As well as translating images from natural to synthetic-looking,
\metvae{} can perform the inverse translation from synthetic to natural-looking as well.
We provide an example in \autoref{fig:syntonat}. 

\subsection{Baselines}
\label{sec:baselines}
We implement two different methods to serve as baselines for our evaluation.

The first is \metvae{} which already outperforms classical baselines like \emph{EfficientNet}~\cite{Tan2019} and \metcv{}~\cite{Watershed2013}. We compare our new \metvaetran{} to \metvae{} and BigTransfer, commonly shortened to \metbit{} \cite{DBLP:journals/corr/abs-1912-11370}. \metvae{} is a previous work of ours upon which \metvaetran{} and the variation \dmetvaetran{} build. It consists of the same architecture but is trained upon a single dataset consisting of natural and synthetic images. Furthermore, less extensive hyperparameter tuning was performed on \metvae{} due to longer training times.\\
The second method we compare to is a transfer learning pipeline from Kolsenikov et al. called \metbit{} that produces state-of-the-art classification results on Cifar-100 and similar datasets in the few shot case (1-10 examples per class). \metbit{} consists of the classical ResNet \cite{DBLP:journals/corr/HeZRS15} architecture but with very long pre-training times on large image corporas and a custom hyperrule that determines the training time and learning rate during transfer dependent on the new dataset size.
Since the data augmentation applied during \metbit{} is not immanently applicable in the cell counting case, we used the same data augmentation as in our own method.
We tested changes to the hyperrule presented in \metbit{} but did not find any significant improvements, therefore used the values provided by the authors.

\section{Transfer learning methodology}
In this chapter, we will describe various experiments to determine which transfer methodology is suited best to the \metvae{} architecture and the final transfer learning methodology used.
\subsection{Transfer Method}

\begin{figure*}[t]
\centering
\includegraphics[width=1\linewidth]{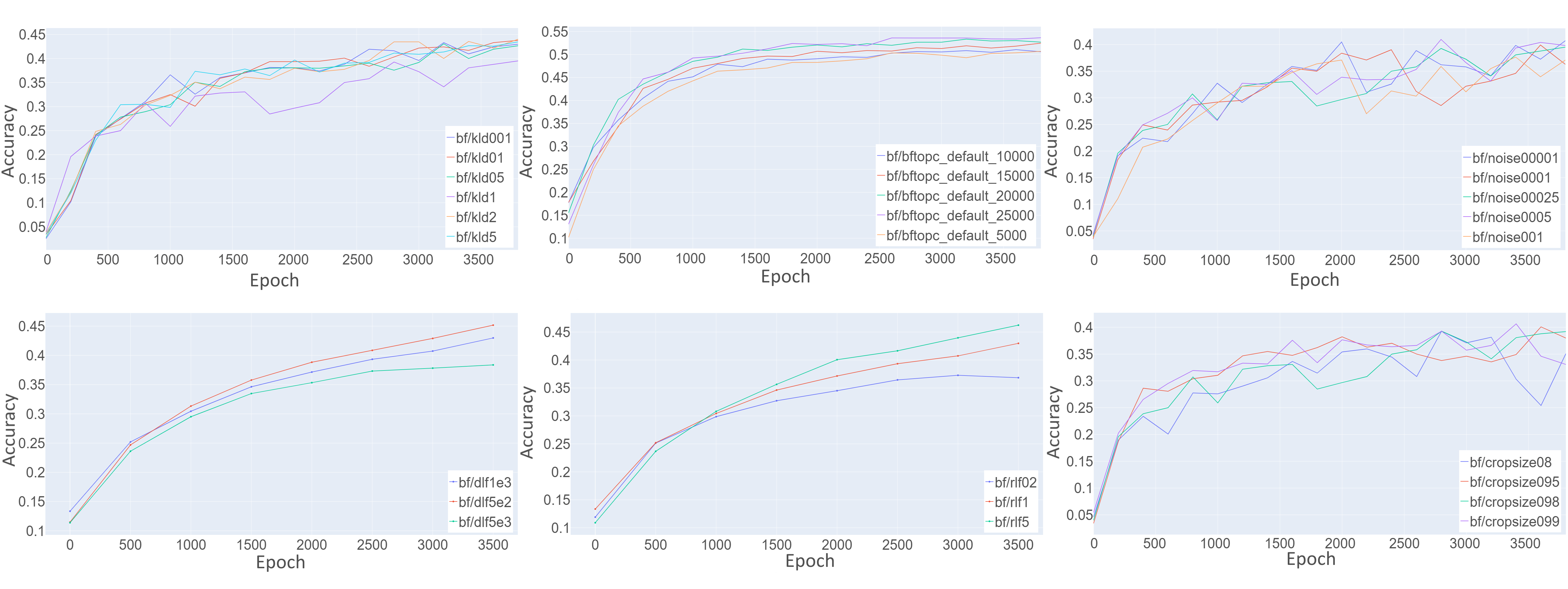}
\caption{Comparison of a variety of hyperparameter choices. These choices include: KLD loss factor (top left), rlf loss factor (bottom middle), dlf loss factor (bottom left), crop size (bottom right), noise factor (top right), length of pre-train (top middle)}\label{fig:hyper}
\end{figure*}

When we write about freezing a section of the network in this work, mathematically, we multiply the gradient update $\Delta$ of the frozen part of the network $f$ with weights $w_f$ by zero. 
Hence, frozen weight update refers to the rule
$ w_f = w_f + \Delta w_f * 0 $ 
instead of the normal weight update
$ w_f = w_f + \Delta w_f $
The gradient is still passed through to non-frozen parts of the network, enabling them to learn.
Since no standard procedure exists in the literature for applying transfer learning to a Twin-VAE which is trained with synthetic data augmentation, four different possible methods of transfer learning are proposed and compared. 
These methods are:

\emph{Frozen Outer layers.} A popular observation in convolutional networks is that the early layers consist of universal edge processing masks and the later layers are more specialized for the task at hand \cite{Yamashita2018}. In a Twin-VAE architecture, these later layers of a standard convolutional network correspond to the shared inner layers and the early convolutional layers correspond to the outer layers of the Twin-VAE. Based on this analogy, we try to train only the inner layers of the network.
Everything which is not part of the shared elements of the network pictured in \autoref{fig:architecture} is not trained.

\emph{Frozen Core.} A common view on VAEs is that the produced embedding space should be highly sensitive in regard to the variance in the training set. Since the imaging method is not changed during normal training the VAE embedding should not encompass this variable, rather it should be highly sensitive to cell count and cell position in the images, which were the main things varied in the original training set. Since the task of cell counting remains the same and the only difference between tasks is the imaging method used, we tried to keep this shared inner representation frozen during training. 
Everything that is part of the shared elements of the network pictured in \autoref{fig:architecture} is not trained.

\emph{Simultaneous transfer.} In the third series of tests, we were not freezing any layers at all. This has the potential problem of the initial transfer period with high losses destroying useful information in intermediate layers.

\emph{Thawing layers.} Last, we experimented to start with frozen inner or outer layers and gradually unfreeze them during training. This is done explicitly to prevent the potential problem described in Simultaneous transfer, but to still be able to fine-tune these layers appropriately to the new task.

\subsection{Hyperparameter tuning}

The original Twin-VAE needed \num{50000} epochs to converge to satisfactory results, which equates to a near \num{100} hours on an NVIDIA Tesla P-100 16G. This made hyperparameter tuning using standard methods computationally costly. 

By using transfer learning to converge significantly faster to similar or even better results, we were able to conduct more extensive hyperparameter searches. 
Since a full grid search over all possible hyperparameters is still not computationally feasible, instead, an iterative search was performed by tuning a single hyperparameter finding the best value and proceeding to the next hyperparameter, recapturing obscured parameter choices in later repetitions.
The hyperparameters and training options tuned were:

\begin{itemize}
\item Training time,
\item transfer method,
\item image noise ratio,
\item image crop size,
\item relative ratio of KLD loss, regression loss and reconstruction loss
\item learning rate
\item learning rate schedule
\end{itemize}
Plots for most of these are provided in \autoref{fig:hyper}. 

\subsection{Results}
We see when tuning the hyperparameters that most of the parameters show only small improvements to the final accuracy (1-2 percent), cumulatively the performance of the network can be significantly improved (5 percent). Our experiments show that for most hyperparameters, large performance degradation can be observed if they are poorly chosen.

The choice of transmission method revealed which parts of the network have learned transferable information and which need to be retrained. We conclude that, unlike in a typical convolutional network, the outer layers need the most retraining. 
If these layers remained frozen, the network could not successfully transfer its knowledge to the new imaging method. Conversely, when the inner layers remained frozen, the network achieved only 1-2 percent less performance than when everything remained unfrozen.

This could be due to the rather unique circumstance of the final task being the same, just on pictures taken with a different imaging method.

Not freezing any layers achieved the best performance overall, we attribute this to the postulated effect of the initial transfer window scrambling information  not being observed. 

\begin{figure}[t]
\centering
\includegraphics[width=\linewidth]{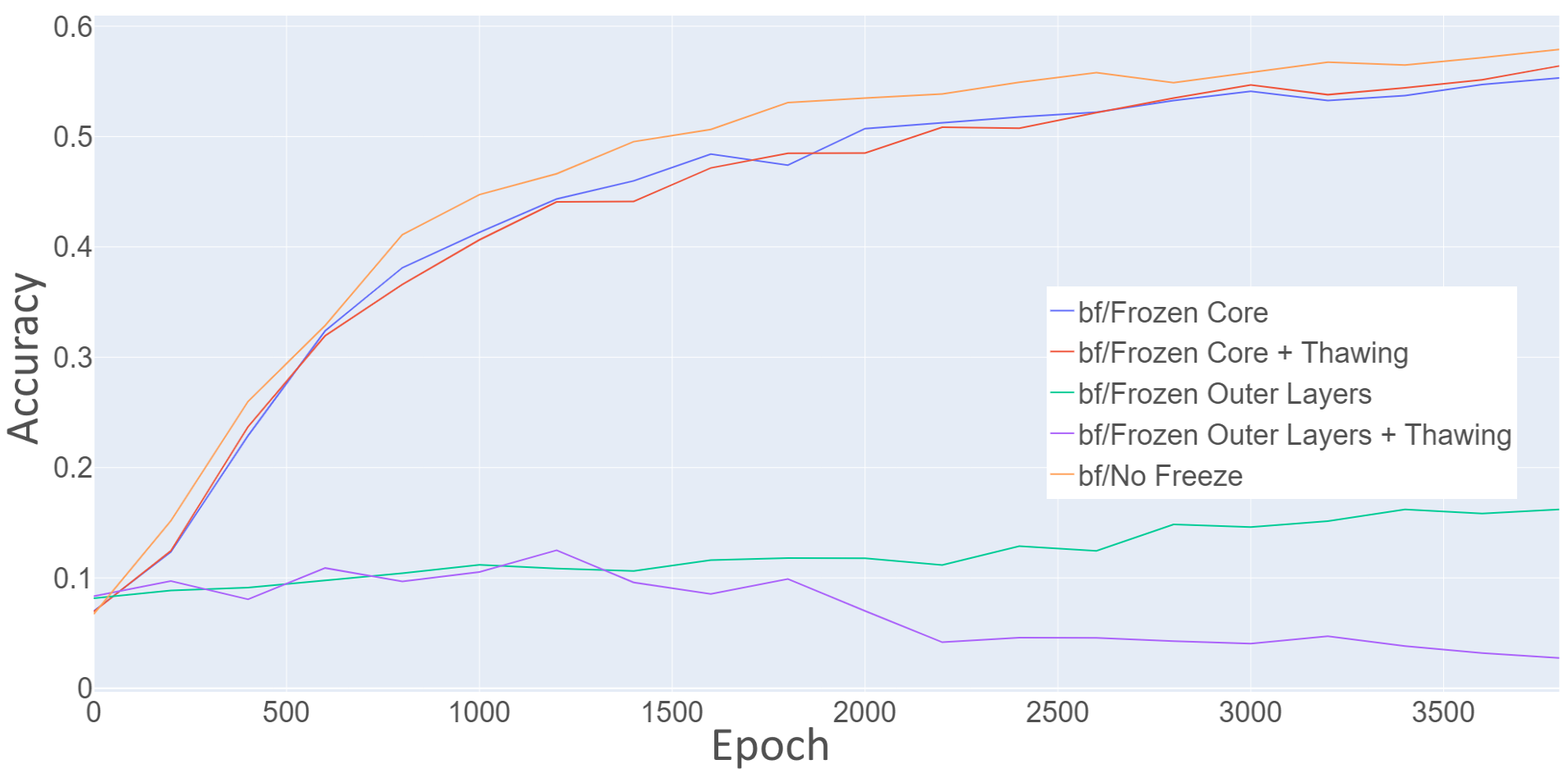}
\caption{Comparison of test accuracy between different transfer learning options. The transfer was performed from the datasets \dssynpclte{}, \dsnatpclte{} transferring to the datasets \dssynbflte{}, and \dsnatbflte{}. In this application, the simplest option (not freezing any layers at all) performs the best, while freezing only the inner layers performs only slightly worse. Freezing the outer layers greatly impact the ability of the network to adapt to the new imaging method. Gradual thawing of the frozen layers does not have a large impact on performance.}\label{fig:testacc}
\end{figure}

Another interesting effect observed was that when trained for very long training times (\num{150000} epochs) the network did not show any signs of double descent \cite{DBLP:journals/corr/abs-1912-02292} and achieved convergence after only \num{10000} epochs. Compared to the non pre-trained network where convergence was achieved at the earliest after \num{50000} epochs, this represents a speed up of at least 5 times.

\section{Twin-VAE during transfer learning}
In this section we systematically investigate the effect the unique architecture of the Twin-VAE has on the transfer process. We choose to investigate whether the decoder part of the network is needed during transfer learning, and whether the synthetic data used for training is needed during transfer learning. Through this we ask if the Twin-VAE architecture is only necessary for the pre-training part and if it can be simplified to a normal convolutional network for transfer learning.
\subsection{Decoder}
\begin{figure}[t]
\centering
\includegraphics[width=\linewidth]{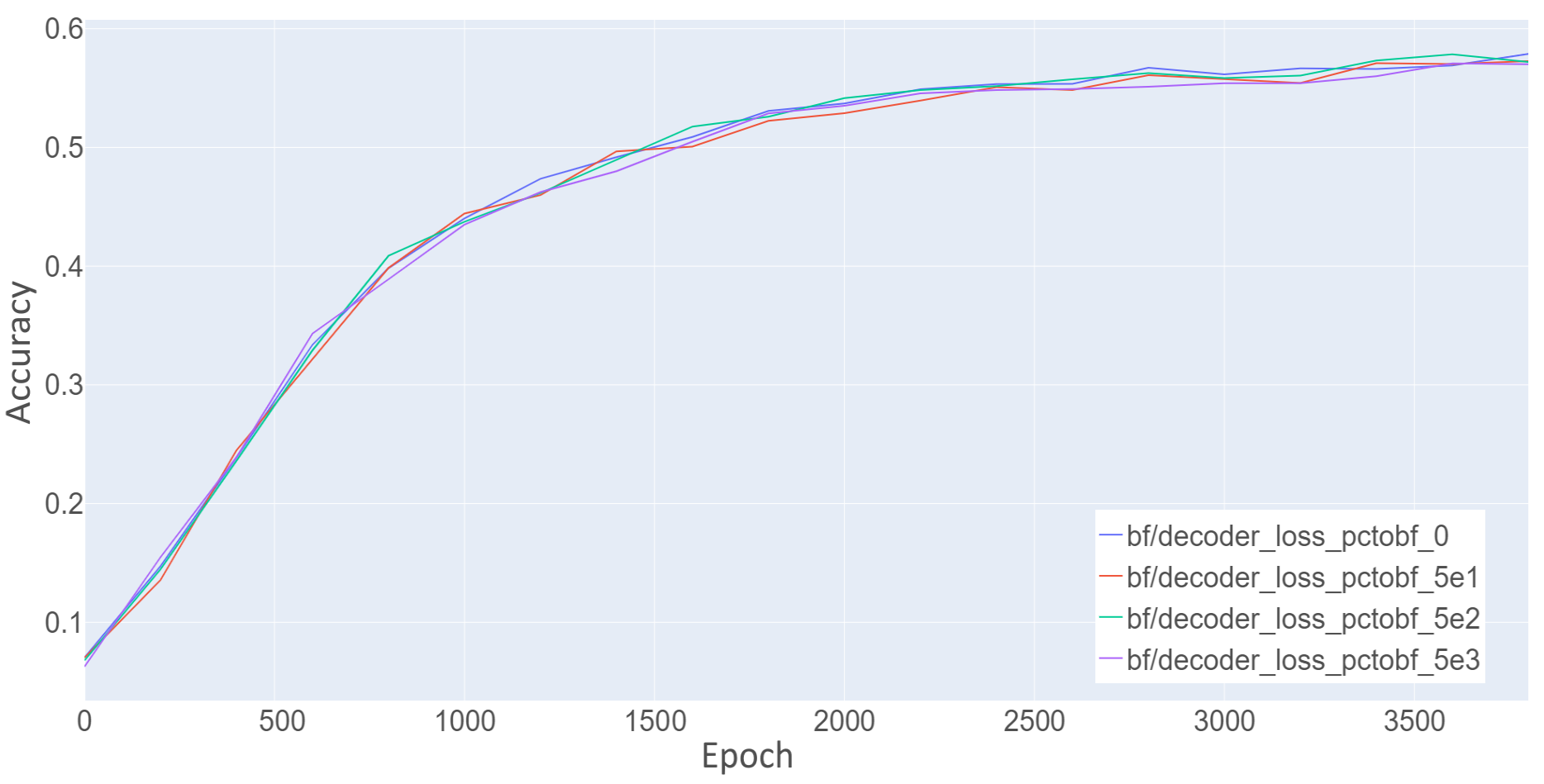}
\caption{Comparison of test accuracy during transfer between different reconstruction loss factors \(\weightparam_{\regucost}^{\imgtype}\). The transfer was performed from the datasets \dssynpclte{}, \dsnatpclte{} transferring to the datasets \dssynbflte{}, and \dsnatbflte{}. All values tried have little to no effect on the accuracy.}\label{fig:decoder}
\end{figure}

To investigate whether the Decoder is important during transfer learning, we perform multiple transfer training runs where we successively lower the reconstruction loss factor \(\weightparam_{\recocost}^{\imgtype}\). \autoref{fig:decoder} shows that the loss factor does not seem to have any impact on performance. The variance between runs is small enough to be within normal statistical deviations observed between runs with the same parameters and can not be clearly attributed to the different loss factors. Based on this, we conclude that the Decoder part of the network does not have a positive impact during the transfer procedure and can be set inactive to speed up transfer computing time even further.

\subsection{Synthetic Data}
To assess the relevance of the synthetic data during transfer learning, we vary the ratio of synthetic data to natural data. In the original paper\cite{TwinVAE} the ratio of synthetic data to natural data was maintained at 1:1 to prevent the architecture from projecting the different data types to distinct embeddings in the VAE bottleneck. A point of note is that a large part of the natural data is unlabeled, while the synthetic data is fully labeled, subsequently the synthetic data had a large contribution to the training of the regressor. 

During our experiments we vary the ratio of synthetic to natural data in the range of 0.25-10:1.

\begin{figure}[t]
\centering
\includegraphics[width=\linewidth]{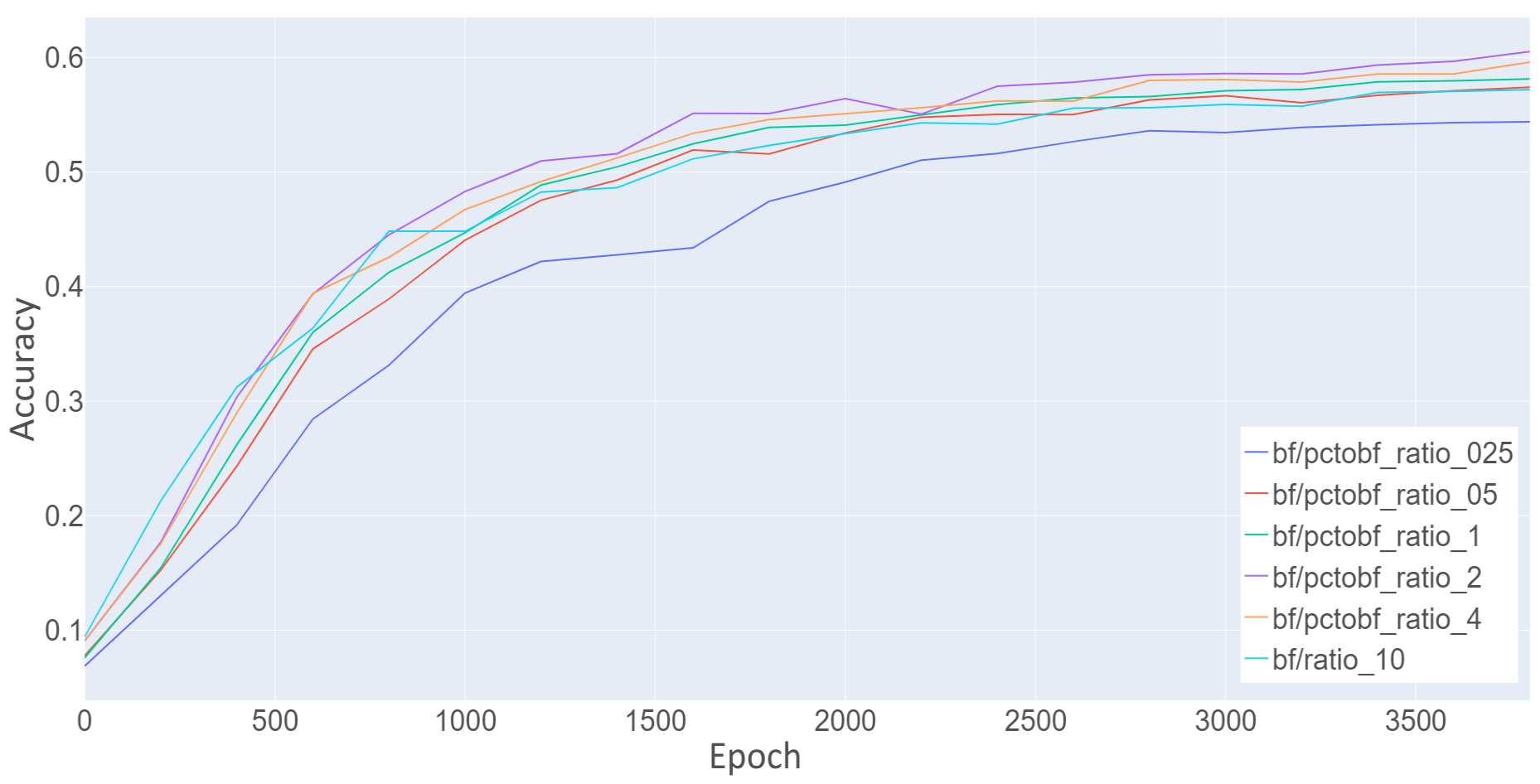}
\caption{Comparison of test accuracy during transfer between different synthetic data to natural data ratios. The transfer was performed from the datasets \dssynpclte{}, \dsnatpclte{} transferring to the datasets \dssynbflte{}, and \dsnatbflte{}. Synthetic data ratio below 0.5:1 negatively impact the network's performance, ratios above 5:1 also have a negative impact upon performance. The best performance was achieved with a ratio of 2:1.}\label{fig:syndata}
\end{figure}

\autoref{fig:syndata} suggests that the performance of the network is negatively affected when the synthetic data ratio is especially small (<0.5) or large (>5). The optimal ratio found was 2:1. Since higher amounts of training data generally lead to better performance, there seems to be a problem generalizing from the synthetic data to the natural data.

\begin{figure}
\centering
\includegraphics[width=\linewidth]{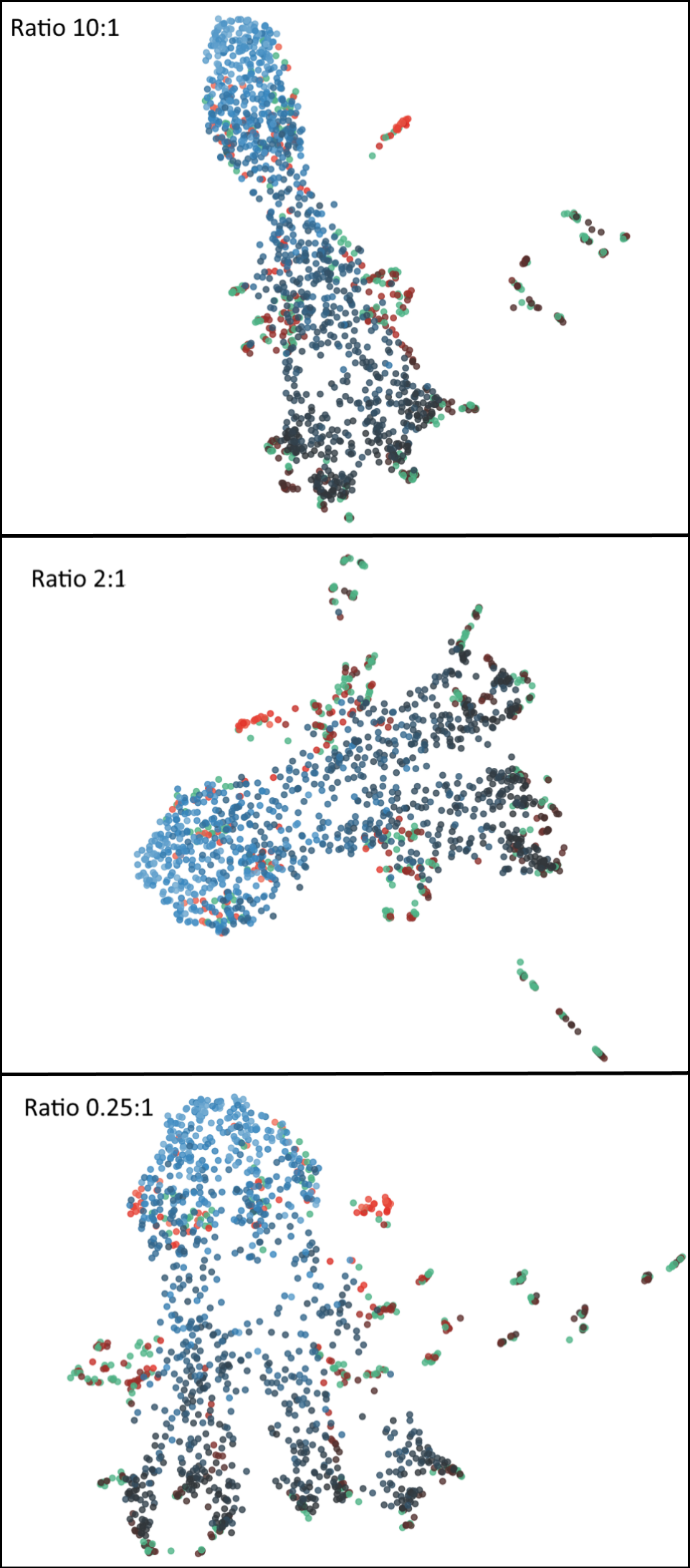}
\caption{Different UMAPs of the embedding dimension of the Twin-VAE after transfer. On the dataset \dssynpclte{}, \dsnatpclte{} transferring to the datasets \dssynbflte{}, and \dsnatbflte{}. Blue dots represent synthetic data, red dots natural data and green dots natural unlabeled data. The color gradient of the dots represent the number of cells in the image, where lighter indicates a higher cell count. In all cases some outliers of red and green dots are found. In the lowest ratio these outliers seem to be most severe. The direction of low to high cell count is clearly recognizable in all plots. The high ratio visualization has the strongest correlation of cell count to embedding distance.}\label{fig:UMAPratio}
\end{figure}

We suggest that the KLD loss \(\regucost\) is still able to keep the distribution of the natural and the synthetic data the same in cases where the ratio is close enough to 1:1 but for more extreme ratios the KLD loss \(\regucost\) alone is insufficient. To validate this hypothesis we show 3 different UMAPs \cite{mcinnes2018uniform} in \autoref{fig:UMAPratio} that depict the distribution of natural and synthetic data in the embedding layer of the VAE.

\autoref{fig:UMAPratio} shows that the VAE distribution seems to regard the number of cells in an image as a more important aspect, the higher the synthetic data ratio. However, it does not show a separation of synthetic (blue dots) and natural data (red dots), so it is not clear why the performance of the model decreases for higher synthetic data ratios. We leave this for future work.

\section{Results and Discussion}
\label{sec:evaluation}
We present the final results of all methods on the four data sets \dssynpclte{}, \dsnatpclte{}, \dssynbflte{}, and\\ \dsnatbflte{} in \autoref{table:results-cell-counts}.
Our \metvaetran{} consistently outperforms all other methods \metvae{} and \metbit{} by a clear margin on the \dssynbflte{}, and \\ \dsnatbflte{} data sets.
On the \dssynpclte{} and \\ \dsnatpclte{} data sets, where more natural data is available for training, the stronger initialization by \metvaetran{} does not have as strong of an impact. It has a good performance on the dataset with very little training time, but does not easily achieve the same performance as \metvae{} on \dsnatpclte{}. On the \dssynpclte{} it outperforms all other methods handily, this is most likely due to the transfer working even better on synthetic images than on natural images. Further reasons for \metvaetran{} not performing better than \metvae{} on \dsnatpclte{} could be that the performance of the initialization on the original dataset is not as high. To remedy this, we used the \metvaetran{} transferred to \dsnatbflte{} and \\ \dssynbflte{} as a starting point to transfer back to \dsnatpclte{} and \dssynpclte{}, using this configuration we obtained the best results on \dsnatpclte{} and \dssynpclte{}, we call this configuration the\\ \dmetvaetran{}.

\begin{table*}[htbp]
\caption{Evaluation of all methods on the data sets \dssynpclte{}, \dsnatpclte{}, \dssynbflte{}, and \dsnatbflte{}.
For each method and data set, we report the mean absolute error (MAE), the mean relative error (MRE), and the accuracy.
Ultimately, only the performance on natural data (\dsnat{}) is important,
but we also report the performance on synthetic data (\dssyn{}) to provide further context.
We use an upward arrow \(\uparrow\) to indicate that higher is better
and a downward arrow \(\downarrow\) to indicate that lower is better. Training times are reported on an NVIDIA Tesla P-100 16G GPU for all models.}
\label{table:results-cell-counts}
\centering
\sisetup{detect-weight,table-auto-round}
\setlength{\tabcolsep}{0.5em} 
\begin{tabular}{cS[table-format=2.2]S[table-format=2.2]S[table-format=2.2]|S[table-format=2.2]S[table-format=2.2]S[table-format=2.2]|S[table-format=2.2]}\toprule
Method & \dssyn{} {MAE  $\downarrow$} & {MRE / \%  $\downarrow$} & {Acc. / \%  $\uparrow$} & \dsnat{}  {MAE $\downarrow$} & {MRE / \%  $\downarrow$} & {Acc. / \%  $\uparrow$} & {training Time / sec $\downarrow$}\\\midrule
\multicolumn{7}{c}{{\dspc{} (phase-contrast microscopy)}} \\
\metvae{} (\dsnat{} only)         &        \na &        \na &        \na &       1.07 &       20.1 &       39.8 & 180000  \\
\metvaemaxacc{}                   & \best  0.09 &       0.68 & 68.2       &       0.60 &       5.92 & \ 57.8 & 400000 \\
\metvaemindev{}                   &       0.14 &       0.73 &       62.1 &  0.59 & \ 5.66 &       57.0 & 400000\\
\metbit{}                               &        \na &        \na &        \na &      2.203 &        \na &      26.13 & 9000 \\
\metvaetran{}(\dsnat{} only) &        \na &        \na &        \na &       1.01 &      14.11 &       44.2 & 24000 \\
\metvaetran{}                &       0.15 &  0.43 & 85.0 &       0.66 &      6.46 &     53.7 &  40000  \\
\dmetvaetran{}                &       0.12 & \best 0.43 & \best 85.0 &       \best0.58 &      \best 5.56 &     \best  58.7 &  71000  \\
\midrule
\multicolumn{7}{c}{{\dsbf{} (bright-field microscopy)}} \\
\metvae{} (\dsnat{} only)          &        \na &        \na &        \na &       0.91 &       13.3 &        23.4  & 150000 \\
\metvaemaxacc{}                    &       0.48 &       4.27 &       60.1 &       0.68 &       7.60 &        53.2  &  310000 \\
\metvaemindev{}                    &       0.52 &       4.63 &       58.2 &       0.63 &       7.31 &        51.9 &  310000 \\
\metbit{}                                &        \na &        \na &        \na &       1.03 &        \na &        43.1 & 5400\\
\metvaetran{}(\dsnat{} only)  &        \na &        \na &        \na &       0.72 &       7.88 &       51.36 & 20000 \\
\metvaetran{}                 & \best 0.40 & \best 3.87 & \best 66.6 & \best 0.52 & \best 5.47 & \best 60.74 &  31000  \\
\bottomrule
\end{tabular}
\end{table*}

The most useful long term information transfer seems to be happening from better performing microscopy methods (phase contrast) to worse performing methods (bright field). In conclusion, a magnitude shorter training times, a better starting point and some hyperparameter tuning always outperforms random weight initialization.
Interestingly, the resulting transfer performance on the \dssynbflte{}, and \dsnatbflte{} datasets is better than either microscopy method alone (\num{60.74} $\leftrightarrow$ \num{53.20}/\num{57.80}).

In summary, the factors gained with our best methodology \dmetvaetran{} are: 19\% better accuracy compared to BiT on the \dsnatbflte{} dataset, 32\% accuracy better accuracy compared to BiT on the \dsnatpclte{} dataset. We achieved even higher accuracy gains compared to EfficientNet and Watershed. We gained about 1.1\% accuracy compared to our previous \metvae{} on the \dsnatpclte{} dataset and about 7.5\% accuracy compared to \metvae{} on the \dsnatbflte{} dataset.

\begin{figure}
\centering
\includegraphics[width=\linewidth]{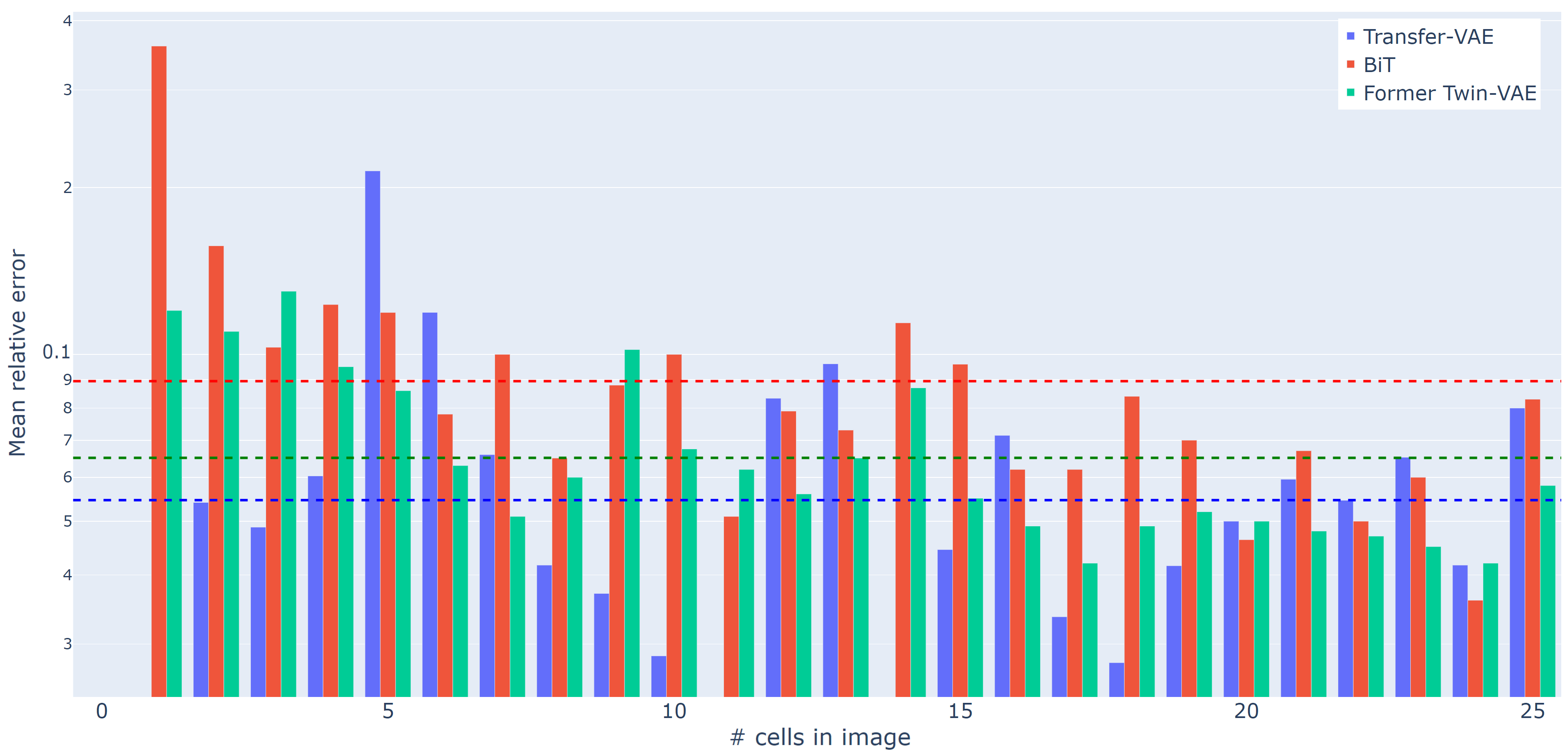}
\caption{\ The mean relative error (MRE) for \metbit{}, \metvaetran{}, and \metvaemindev{} on \dsnatbflte{} on a logarithmic scale. Horizontal bars indicate the average MRE of their respective color and method.}
\label{fig:deviation}
\end{figure}

\section{Conclusion and Outlook}
In this paper, we present a significant improvement over the original Twin-VAE by using transfer learning methods to improve the accuracy and training times of the original architecture using pre-trained checkpoints of the original paper. Utilizing these shortened training times, we perform extensive hyperparameter tuning and improve the accuracy even further.
Furthermore, we research which parts of the original Twin-VAE architecture are necessary for the transfer learning case.
We determine that the synthetic data still plays a key role in achieving high performance and can not be removed without significant performance loss. 
The Decoder part of the network does not contribute to achieving higher accuracies during transfer learning and is therefore only necessary for pre-training on the original datasets.

\emph{Limitations:} 
The transfer procedure only performs well if the initial starting point has a high performance on the original dataset. If the starting point has low performance on the original dataset the transfer procedure might achieve lower performance than a normally trained network.
In practice this is not a big problem since good starting points (in our case the \metvaemaxacc{}) can be chosen easily based upon their performance on their original dataset.

The Twin-VAE architecture can use synthetic data to improve performance on natural data, currently this is limited to ratios of up to 2:1. It would be desirable if the network could abstract even further and possibly not need any labels on the natural data at all. Current methods to increase regularization (Higher \(\regucost\), Dropout, Weight Decay) are not able to force the network to project the different data types onto the same embedding.

\emph{Potential and future work:}
The methods described in this paper enables the automation and remote surveillance of various previously tedious and labor-intensive laboratory experiments. To use its full potential it would be interesting to implement this method as edge computing on modern microscopy hardware.

To enable edge computing computational efficiency is key, here different techniques to prune network weights or similar methods could be explored to facilitate even faster computation times.

Other possible future work includes, using active learning~\cite{Cohn1994} to further refine the algorithm's predictions and enable entirely new areas of prediction, such as the survival probability of an entire cell culture. 






\section{Authors' information}
\label{sec:8}
\textbf{Data Availability}
The datasets generated during and/or analysed during the current study are available in the Natural and synthetic CHO-K1 time-lapse suspension cell microscopy images (bright-field and phase-contrast) v2 repository, \url{https://pub.uni-bielefeld.de/record/2960030}.

\begin{acknowledgements}
\thanks{We gratefully acknowledge funding by the BMWi witin the project KI-Marktplatz, grant number 01MK20007E (PK), and by the European Commission within the project  ICU4Covid, Grant Agreement number 101016000 —  H2020-SC1-PHE-CORONAVIRUS-2020-2 / H2020-SC1-PHE- CORONAVIRUS-2020-2-CNECT (DS).}
\end{acknowledgements}

\section*{Conflict of interest}
The authors declare that they have no conflict of interest.

\bibliographystyle{spmpsci} 
\bibliography{main} 

\end{document}